\documentclass[sigconf]{acmart}

\usepackage{booktabs} 

\usepackage[inline]{enumitem}
\usepackage{amsmath}
\usepackage{graphicx, subcaption}

\usepackage{soul} 
\usepackage{xcolor} 

\copyrightyear{2019}
\acmYear{2019} 
\setcopyright{iw3c2w3}
\acmConference[WWW '19]{Proceedings of the 2019 World Wide Web Conference}{May 13--17, 2019}{San Francisco, CA, USA}
\acmBooktitle{Proceedings of the 2019 World Wide Web Conference (WWW '19), May 13--17, 2019, San Francisco, CA, USA}
\acmPrice{}
\acmDOI{10.1145/3308558.3313559}
\acmISBN{978-1-4503-6674-8/19/05}

\usepackage{balance}

\begin{document}
\title{Fairness in Algorithmic Decision Making:\\ An Excursion Through the Lens of Causality}


\author{Aria Khademi}
\affiliation{%
  \institution{Pennsylvania State University}
}
\email{khademi@psu.edu}

\author{Sanghack Lee}
\affiliation{%
  \institution{Purdue University}
  }
\email{lee2995@purdue.edu}

\author{David Foley}
\affiliation{%
  \institution{Pennsylvania State University}
  }
\email{djf47@psu.edu}

\author{Vasant Honavar}
\affiliation{%
  \institution{Pennsylvania State University}
  }
\email{vhonavar@psu.edu}

\begin{abstract}
As virtually all aspects of our lives are increasingly impacted by algorithmic decision making systems, it is incumbent upon us as a society to ensure such systems do not become instruments of unfair discrimination on the basis of gender, race, ethnicity, religion, etc. We consider the problem of determining whether the decisions made by such systems are discriminatory, through the lens of causal models. We introduce two definitions of group fairness grounded in causality: {\em fair on average causal effect} (FACE), and {\em fair on average causal effect on the treated} (FACT). We use the Rubin-Neyman {\em potential outcomes} framework for the analysis of cause-effect relationships to robustly estimate FACE and FACT. We demonstrate the effectiveness of our proposed approach on synthetic data. Our analyses of two real-world data sets, the Adult income data set from the UCI repository (with gender as the protected attribute), and the NYC Stop and Frisk data set (with race as the protected attribute), show that the evidence of discrimination obtained by FACE and FACT, or lack thereof, is often in agreement with the findings from other studies. We further show that FACT, being somewhat more nuanced compared to FACE, can yield findings of discrimination that differ from those obtained using FACE.
\end{abstract}

\maketitle

\section{Introduction} \label{intro}
With the growing adoption of algorithmic decision making systems, e.g., AI and machine learning systems, across many real-world decision making scenarios on the Web and elsewhere, there is a pressing need to make sure that such systems do not become vehicles of unfair discrimination, inequality, and social injustice \cite{barocas2016big,barocas2017big}. Of particular interest in this context is the task of detecting and preventing discrimination or unfair treatment of individuals or groups on the basis of gender, race, religion, etc. Such discrimination is traditionally addressed using one of two legal frameworks: {\em disparate treatment} (which aims to enforce procedural fairness, namely, the equality of treatment that prohibits the use of the protected attribute in the decision process); and {\em disparate impact} \cite{barocas2016big} (which aims to guarantee outcome fairness, namely, the equality of outcomes between protected groups relative to other groups). It is clear that enforcing procedural fairness within the disparate treatment framework does not guarantee non-discrimination within the disparate impact framework.

There is growing interest in algorithmic decision making systems that are demonstrably {\em fair} (see \cite{berk2017fairness} for a review). Much of this literature relies on precise definitions that quantify fairness to avoid discrimination with respect to {\em protected} attributes, 
e.g., race, gender, on the basis of the legal notions of disparate treatment or disparate impact \cite{barocas2016big} (see \cite{romei2014multidisciplinary,zliobaite2015survey,barocas2016big,berk2017fairness,loftus2018causal} for reviews). Some examples include: fairness through unawareness \cite{grgic2016case}, 
individual fairness \cite{dwork2012fairness}, 
equalized odds \cite{hardt2016equality,zafar2017fairness_www}, 
calibration \cite{chouldechova2017fair}, 
demographic (or statistical) parity \cite{calders2009building,kamishima2012fairness,kamiran2009classifying,johndrow2017algorithm}, 
the 80\% rule (disparate impact) \cite{feldman2015certifying,zafar2017fairnes_aistats}, 
representational fairness \cite{zemel2013learning,louizos2015variational}, 
and fairness under composition \cite{dwork2018fairness}.

Unfortunately, choosing the appropriate definition of fairness in a given context is extremely challenging due to a number of reasons. First, depending on the relationship between a protected attribute and data, enforcing certain definitions of fairness can actually increase discrimination \cite{kusner2017counterfactual}. Second, different definitions of fairness can be impossible to satisfy simultaneously \cite{kleinberg2016inherent,berk2017fairness,chouldechova2017fair}. 
Many of these difficulties can be attributed to the fact that fairness criteria are based \emph{solely} on the joint probability distribution of the random variables of interest, namely, \(\hat{Y}\) (predicted outcome), \(Y\) (actual outcome), \(\tilde{X}\) (features), and \(A\) (sensitive attributes). \cite{hardt2016equality} recently showed any such definition for fairness of a predictor that depends merely on the joint probability distribution is not necessarily capable of detecting discrimination. Hence, it is tempting to approach the problem of fairness through the lens of causality \cite{barabas2018interventions}. 

Answering questions of fairness through the lens of causality entails replacing the question ``Is the decision discriminatory with respect to a protected attribute?'' by: ``Does the protected attribute have a causal effect on the decision?'' A practical difficulty in using this approach is that, in general, establishing a causal relationship between a protected attribute and a decision requires the results of experimental manipulation of the protected attribute. Fortunately, however, existing frameworks for determining causal effects from observational data \cite{pearl2009causality,imbens2015causal} provide a rich set of theoretical results as well as practical tools for elucidating causal effects, and specifically, answering questions about {\em counterfactuals} or {\em potential outcomes}, i.e., results of {\em hypothetical} experimental interventions from observational data, whenever it is possible to do so. Hence, there is a growing body of work (see \cite{loftus2018causal} for a recent review) focused on explicitly causal (as opposed to purely joint distribution based or {\em observational}) definitions for fairness (e.g., \cite{kusner2017counterfactual,kusner2018causal,zhang2018fairness,nabi2018fair,kilbertus2017avoiding,chiappa2018path,zhang2017causal,bonchi2017exposing,li2017discrimination,zhang2016situation,russell2017worlds,vanderweele2014causal}). While some, e.g., \cite{zhang2017causal}, have focused on testing fairness (or conversely, determining whether there is discrimination), others, e.g., \cite{kilbertus2017avoiding,kusner2017counterfactual} have sought to design machine learning algorithms that yield predictive models that are demonstrably fair. However, most of the existing work on defining fairness in causal terms has focused on variants of individual fairness. Against this background, we focus on robust methods for detecting and quantifying discrimination against protected groups, which is a necessary prerequisite for developing predictive models that are provably non-discriminatory.


\subsubsection*{\bf Contributions.}
We reduce the problem of quantifying discrimination against protected groups to the well-studied problem of estimating the causal effect of some variable(s) on a target (outcome) variable. We introduce two explicitly causal definition of fairness in a population, {\em fair on average causal effect} (FACE), and in a protected group, {\em fair on average causal effect on the treated} (FACT), both with respect to a protected attribute (e.g., gender, race). We use the Rubin-Neyman {\em potential outcomes} framework \cite{rubin1974estimating,rubin2005causal,imbens2015causal} for robust estimation of FACE and FACT. We demonstrate the effectiveness of the proposed approach in detecting and quantifying group fairness using synthetic data, as well as two real-world data sets: the Adult income data from the UCI repository \cite{Dua:2017} (with gender being the protected attribute), and the NYC Stop and Frisk data (with race being the protected attribute). We show that the evidence of discrimination, or lack thereof, obtained by FACE and FACT is often in agreement with other studies. We further show that FACT,  being somewhat more nuanced compared to FACE, can yield findings of discrimination that differ from those obtained using FACE.

\section{Fairness: A Causal Perspective} \label{causal_fairness}
Assume we have observational data on a population of individuals. Let \(\tilde{X} \in \mathcal{X}\) be the vector of non-protected attributes, \(A \in \mathcal{A} = \{a, a^\prime \}\) be a binary protected attribute, and \(Y \in \mathcal{Y}\) an outcome of interest. The question we want to answer is: Are individuals being discriminated against, on average, with respect to outcomes or decisions \(Y\) on the basis of a protected attribute \(A\)? From a causal perspective, such a question is \emph{equivalent} to the following question: Does \(A\) have a causal effect on \(Y\)? In other words, how much would \(Y\) change, on average, were the value of \(A\) to change? Both Structural Causal Models \cite{pearl2009causality} and the Rubin-Neyman Causal Model (RCM) \cite{imbens2015causal} (also called the \emph{potential outcomes} model) offer methods for estimating such causal effects from observational data.

We introduce two explicitly causal definitions for fairness ``on average'' in a population or a protected group (as opposed to causal definitions of individual fairness, e.g., counterfactual fairness \cite{kusner2017counterfactual}) with respect to a protected attribute (e.g., gender, race). Let \(Y_i^{(a)}\) and \(Y_i^{(a^\prime)}\) be the potential outcomes of a data point \(i\) had their value of \(A\) been \(a\) and \(a^\prime\), respectively. Let \(h: \mathcal{X} \times \mathcal{A} \to \mathcal{Y}\) be a decision function (or a predictive model trained using machine learning) that is used to support decision making. $\mathbb{E}[\cdot]$ is the expectation of a random variable. We define the following.

\begin{definition}\label{ave_fair}(FACE: Fair on Average Causal Effect). 
	A decision function \(h\) is said to be fair, on average over all individuals in the population, with respect to \(A\), if $\mathbb{E}[Y_i^{(a)} - Y_i^{(a^\prime)}] = 0$.
\end{definition}

\begin{definition}(FACT: Fair on Average Causal Effect on the Treated).
	 A decision function \(h\) is said to be fair with respect to \(A\), on average over individuals with the same value of \(A\), if \(\mathbb{E}[Y_i^{(a)} - Y_i^{(a^\prime)} \, | \, A_i = a] = 0\).
\end{definition}


\subsubsection*{{\bf Example}} Imagine we are given the hiring data of a company containing demographic information about applicants, as well as \(A = \) \{male, female\} as their gender, and \(Y=\) \{hired, rejected\} as whether they were hired by the company. Our task is to determine whether the company's hiring decisions are fair on average with respect to gender. FACE contrasts the expected outcomes (i.e., hiring) between men vs. women with the expectation taken over the entire population. FACT contrasts the expected outcomes observed for a specific protected group (e.g., women) and the hypothetical (counterfactually inferred) outcomes for the group had they not been members of the protected group (with the expectation taken only over the members of the protected group), e.g., hiring outcomes for women contrasted with outcomes for the same individuals had their gender been different with all other attributes remaining unchanged. Obviously, such counterfactual outcomes cannot be obtained from observational data\footnote{This is called the Fundamental Problem of Causal Inference (FPCI) from observational data \cite{holland1986statistics}.} and ought to be estimated. 

\section{Estimating FACE and FACT} \label{estimate_act}
We use tools offered by the potential outcomes framework \cite{imbens2015causal} to estimate FACE and FACT. These tools rely on the following key assumptions: i) \textit{Consistency} which requires that for a data point \(i\), the potential outcome of \(i\) under any level of treatment \(a\), i.e., \(Y^{a}_i\), equals the \emph{actual} outcome observed for that data point, \(Y^{obs}_i\), had they been exposed to treatment \(a\). Formally, under consistency, \(Y_{i}^{obs} = a \, Y_{i}^{(a)}+ a^\prime \, Y_{i}^{(a^\prime)}\) would hold for all \(i\). This assumption, used in existing literature \cite{nabi2018fair,chiappa2018path,pearl2019manipulation,madras2019fairness}, is a rather natural one to make in our setting. ii) \textit{Positivity} which asserts that the probability 
\(Pr(A=a \, | \, X=x)>0\) for all values of \(A\). In our setting, this means each value of the protected attribute has a non-zero probability. iii) \textit{Stable Unit Treatment Value Assumption (SUTVA)} \cite{rubin1980randomization} which consists of two sub-assumptions:
	1) Absence of {\em interference} between individuals \cite{cox1958planning}, which means that an individual's potential outcome is unaffected by the treatment assigned to any other individual. 
	While this assumption is plausible in our setting, it may be violated in some settings, in which case, such violations should be accounted for \cite{robins2018causality}.
	2) Presence of only one form of treatment (and control). For example, if a treatment involves administering a drug, then all individuals who take the drug, take it in the same form (e.g., injection). This assumption is trivially satisfied in our setting because treatment is simulated by the protected attribute.
iv) \textit{Unconfoundedness of the treatment mechanism} which implies that given a set of observables, the potential outcomes of each individual are jointly independent of the corresponding treatment \cite{rubin1978bayesian}. Unconfoundedness cannot be verified or contradicted entirely on the basis of observational data. However, sensitivity analysis \cite{rosenbaum2005sensitivity,liu2013introduction} can be a useful tool for analyzing the estimated causal effects under violations of the unconfoundedness assumption. 
Strong ignorability refers to the combination of unconfoundedness and positivity \cite{rosenbaum1983central}. 
Strong ignorability is a sufficient condition for the causal effect to be identifiable \cite{robins2018causality} and is equivalent to the {\em back-door criterion} \cite{pearl2010foundations}, which is required for identifiability of the causal effects in Pearl's model of causality \cite{pearl2010foundations}. 
In our work, as in the case of existing work on causal definitions of fairness  \cite{nabi2018fair}, we assume strong ignorability.

\subsection{Estimating and Interpreting FACE}
We use Inverse Probability Weighting (IPW), also known as Inverse Probability of Treatment Weighting (IPTW) in Marginal Structural Models (MSM) \cite{robins2000marginal} to estimate FACE. Specifically, for each individual \(i\), we calculate a \textit{stabilized weight}: $sw_i = \frac{Pr(A_i = a)}{Pr(A_i = a \, | \, \tilde{X}_i = \tilde{x_i})}$ (call it the weight model). 
We obtained stabilized weights using the R package \emph{ipw} (version 1.0-11) \cite{van2011ipw}. Assigning such a weight to every data point, we generate a ``pseudo-population'' in which there are \(sw_i\) copies of each data point \(i\). Subsequently, the associative parameter \(\beta\) in the weighted regression (call it the outcome model) of the (continuous) outcome \(Y\) on the protected attribute \(A\): $\mathbb{E}[Y^{(A)}] = \delta + \beta A + \tilde{\theta}^\top \tilde{X}$,
would be the causal effect of \(A\) on \(Y\). For a binary output \(Y\), we use the weighted logistic regression model: $\mathrm{logit} \, (\mathbb{E}[Y^{(A)}]) = \delta + \beta A + \tilde{\theta}^\top \tilde{X}.$ In the absence of unmeasured confounders, if \emph{either} the weight model 
\emph{or} the outcome model 
are correctly specified, 
then \(\hat{\beta}\) 
is an unbiased estimator of the average causal effect \cite{robins2000marginal}. For example, suppose \(Y\) is salary and \(A\) is gender. 
At the chosen level of statistical significance \(\alpha\), \(\hat{\beta}=0\) implies that salary is fair with respect to gender on average over the entire population of individuals; \(\hat{\beta} \neq 0 \) implies that, on average, women's salary differs from that of men by a factor of \(\hat{\beta}\) (across the entire population). For a continuous outcome \(Y\), 
\(\hat{\beta}\) is simply the average causal effect of \(A\) on \(Y\). For a binary outcome \(Y\), \(\hat{\beta}\) corresponds to the causal odds ratio of salary for women versus men. 



\subsection{Estimating and Interpreting FACT}
We use \emph{matching} to estimate FACT. Consider the example of salary discrimination based on gender. For a woman, we can never observe what the salary would have been, \emph{had she been a man} (i.e., her counterfactual salary). Hence, we estimate the counterfactual salary as follows \cite{imbens2015causal}: 
1) Using a suitable matching technique (see Section Matching Methods), we match the woman \(i\), to a man \(j\) who is closest to \(i\) with respect to a distance measure \(d(i,j)\). 2) The matching process is repeated as needed until matches are of acceptable quality (see Section Quality of Matches). 3) After matching, we use the salary of the matched man \(j\) (i.e., \(Y_j\)), as the counterfactual salary of the woman \(i\). 

\subsubsection*{{\bf Matching Methods}} \label{matching_algo}
The results of matching depend on the choice of distance measure \(d(\cdot,\cdot)\) as well as the matching process. Several matching methods exist (see \cite{stuart2010matching} for a survey). In what follows, for simplicity and brevity, we refer to individuals with protected attribute set to \(A=a\) as the \emph{treated} individuals and those with the protected attribute set to \(A = a^\prime\) as the \emph{controlled} individuals. We used the matching methods implemented within the R package \emph{MatchIt} (version 3.0.2) \cite{ho2011matchit} with all parameters set to their default values unless otherwise noted:
(i) Exact Matching (EM); 
(ii) Nearest Neighbor Matching (NNM) with propensity score \cite{rosenbaum1983central}. 
Following \cite{rubin2001using}, we estimated the propensity scores using the logit link and transformed them to the linear scale. Then, we ran NNM with replacement, based on the linear propensity scores, and discarded the data points (both from treated and controlled) that fall outside the support of the distance measure;
(iii) Nearest Neighbor Matching with a Propensity Caliper (NNMPC)\label{nnmpc}.  NNMPC includes only matches within a certain number of standard deviations of the distance measure and discards the rest. In NNMPC, we use the same procedure as in NNM, augmented with a caliper = 0.25 \cite{rosenbaum1985constructing}, resulting in the matches outside 0.25 times the standard deviation of the (transformed) linear propensity score, being discarded;
(iv) Mahalanobis Metric Matching within the Propensity Caliper \cite{rubin2001using} (MMMPC). 
MMMPC  determines for each data point, a ``donor pool'' of available matches within the propensity caliper. Mahalanobis metric matching is then performed among the data points chosen in the previous step mimicking blocking in randomized experiments \cite{rubin2001using}. We ran MMMPC with caliper, replacement, and discarding strategy as described above in NNM;
and (v) Full Matching (FM) 
\cite{rosenbaum1991characterization}. We used the same distance measure and discarding strategy as described above in NNM.

\subsubsection*{{\bf Quality of Matches}} \label{match_quality}
To ensure accurate estimation of FACT, it is crucial to measure the ``goodness-of-match.'' If the data points are well matched, then one can proceed to estimate FACT. Common diagnostics for examining the quality of match include both numerical and graphical criteria. Among the numerical criteria, following \cite{rubin2001using}, we compare the standardized difference in the means of the treated and the controlled data points in terms of the distance measure. We denote the absolute value of this difference in means on the original, and matched data, by \(\overline{D}_{a, a^\prime}\), and \(\overline{D}^{m}_{a, a^\prime}\), respectively. For the match to be of good quality, \(\overline{D}^{m}_{a, a^\prime}\) has to be close to \(0\). Among the graphical criteria, we use quantile-quantile (QQ), and jitter plots recommended by \cite{stuart2010matching,ho2011matchit}.\footnote{We avoid the commonly used hypothesis tests for \emph{assessing feature balance} in diagnosing the quality of matches because such tests have been shown to be misleading in general \cite{ImaKinStu08}.}

\subsubsection*{{\bf Outcome Analysis After Matching}} \label{outcome_analysis}
With good quality matched pairs identified, we can proceed to conduct outcome analysis for FACT estimation. Matching methods often assign appropriate weights to the matched data points to balance the treated and controlled data distributions. After obtaining the weights via matching, 
we run the following weighted regression models:
	$\mathbb{E}[Y^{(A)}] = \delta + \gamma A + \tilde{\theta}^\top \tilde{X}$,
for continuous, and
$\mathrm{logit} \, (\mathbb{E}[Y^{(A)}]) = \delta + \gamma A + \tilde{\theta}^\top \tilde{X}$,
for binary outcomes, 
both \emph{on the matched data set}, to estimate FACT. The estimated coefficient for \(A\) in the equations above, i.e., \(\hat{\gamma}\), estimates FACT. The resulting estimate is ``doubly robust'' in that if \emph{either} the matching model, \emph{or} the outcome model, are correctly specified, \(\hat{\gamma}\) would be statistically consistent \cite{ho2011matchit}.

\subsubsection*{{\bf Interpreting \(\hat{\gamma}\) as a Measure of FACT}} \label{interpret_act}
Suppose \(Y\) is salary and \(A\) is gender. 
At the chosen level of statistical significance \(\alpha\), \(\hat{\gamma}=0\) implies that there is no significant difference in expected salary for women compared to what their salary would have been had they been men (with all non-protected attributes \(\tilde{X}\) remaining unchanged, a condition that is approximated by counterfactual inference using matching), thus implying no gender-based discrimination in salary for women; \(\hat{\gamma} \neq 0 \) implies that, on average, women's salary is statistically significantly different from what it would have been, had they been men, thus implying gender-based discrimination in salary. For a continuous outcome \(Y\), e.g., the salary in US dollars, if statistically significant, \(\hat{\gamma} \neq 0 \) means that on average, \emph{considering men and women that are matched based on their feature vector \(\tilde{X}\)}, the difference between women's salary and that of men is \(\hat{\gamma}\). For a binary outcome, e.g., salaries binarized with an arbitrary threshold \(\tau\), \(\hat{\gamma}\) is the causal odds ratio of women's salary compared to that of men, for those women and men who are similar. 

\subsubsection*{{\bf Impact of Unmeasured Confounders on \(\hat{\gamma}\)}}
What if the strong ignorability assumption (i.e., no hidden confounders) is violated? In the absence of unmeasured confounding, matching estimators are unbiased if the matching model is specified correctly, i.e., if balance is achieved over the \emph{observed attributes}. However, it is conceivable that the results of matching could change in the presence of \emph{unobserved confounders} (i.e., hidden bias). We perform sensitivity analysis \cite{rosenbaum2005sensitivity,liu2013introduction} to investigate the degree to which the unmeasured confounders impact \(\hat{\gamma}\). Let \(\Gamma\) be the odds ratio of matched (using any matching method) data points \(i\) and \(j\) receiving a treatment. Sensitivity analysis proceeds by first assuming \(\Gamma = 1\) (i.e., no hidden bias). Then, it increases \(\Gamma\) (e.g., \(1, \ldots, 5\)), thus mimicking the presence of hidden bias, and examines the resulting changes to statistical significance of \(\hat{\gamma}\). The \(\Gamma\) at which the significance of the upper bound for the p-value would change (e.g., from \(< 0.05\) to \(> 0.05\)) is the point at which \(\hat{\gamma}\) is no longer robust to hidden bias. We ran sensitivity analysis using the R package \emph{rbounds} (version 2.1) \cite{keele2010overview}.

\section{Experiments and Results}
We tested our approach on a synthetic data set (where the discrimination based on a protected attribute can be varied in a controlled fashion), and two real-world data sets that have been previously used in studies of fairness. 
In each case, we designated a protected attribute and estimated FACE and FACT as measures of discrimination based on that attribute. We run all of our statistical significance tests with \(\alpha = 0.05\). We proceed to describe the data sets, experiments, as well as our FACE and FACT analyses in detail.
\subsection{Data sets}
\subsubsection*{{\bf Synthetic data set}}\label{synth}
We generated 1000 data points, each with a feature vector \(\tilde{X} = (X_1,\ldots,X_5)\), a protected attribute \(A \in \{0,1\}\), and an outcome variable \(Y\) according to the following:
$X_1,\ldots,X_5 \overset{iid}{\sim} \mathcal{N}(0,1)$; 
$A \, | \, X \sim \mathrm{Bernoulli} \, (\mathrm{logit}^{-1}(\sum_{i=1}^{5} X_i))$;
$Y \, | \, X, A = \sum_{i=1}^{5} X_iW_i$,
where \(\tilde{W} = (W_1,\ldots,W_5)\) is a weight vector (fixed for all data points) with each element drawn randomly in [0,1]. The resulting generative model ensures there are no hidden confounders and there is no discrimination, as measured by FACE and FACT, with respect to the outcome variable \(Y\) on the basis of the protected attribute \(A\). 
\subsubsection*{{\bf The Adult data set}}
The Adult income data set \cite{kohavi-nbtree}\footnote{\url{https://archive.ics.uci.edu/ml/datasets/adult}}, contains information about individuals as well as their salaries. The data set includes 48842 individuals each with 14 attributes, 6 continuous and 8 categorical, including demographic and work-related information such as age, gender, hours of work per week, etc. We examined whether there is gender-based discrimination in salaries 
by designating \emph{gender} as the sensitive attribute. We encoded categorical variables using one-hot-encoding and removed data records with missing values, yielding a data set with 46033 individuals and 45 features (excluding gender, the protected attribute). We designated the outcome \(Y\) to be a binary variable denoting whether the person's annual salary is \(>\$50K\) (Y=1), or \(\leq \$50K\) (Y= 0).

\subsubsection*{{\bf The NYC Stop and Frisk (NYCSF) data set}}
We retrieved the publicly available stop, search, and frisk data from The New York Police Department (NYPD)\footnote{\url{https://www1.nyc.gov/site/nypd/stats/reports-analysis/stopfrisk.page}} website which serves demographic and other information about drivers stopped by the NYC police force. Our question is whether the arrests made after stops have been discriminatory with respect to race. Following \cite{kusner2017counterfactual}, we restricted our experiment to the year 2014 yielding a total of 45787 records. We selected the subset of records corresponding to only Black-Hispanic and White men. We designated \emph{race} as the protected attribute with \(A=1\) denoting Black-Hispanic and \(A=0\) denoting White. We dropped the data records with missing values and encoded categorical variables with one-hot-encoding. The resulting data consist of 7593 records each with 73 features (excluding race, the sensitive attribute). The outcome \(Y\) denotes whether an arrest was made (\(Y=1\)), or not (\(Y=0\)).


\subsection{FACE Check: Fairness Analysis Using FACE}\label{face}
We report our analysis of fairness using FACE, for the synthetic, Adult, and NYCSF data sets. The estimated FACE (\(\hat{\beta}\)) are shown in Table \ref{table:ace}. In all cases, the null hypothesis is \(H_0:\beta = 0\).
In the case of synthetic data, we find insufficient evidence to  
reject \(H_0\), suggesting the outcome is fair with respect to the protected attribute (an expected conclusion given the design of the generative model in Section \ref{synth}, which ensures that the outcome is fair with respect to the protected attribute). 
In the case of Adult data, we reject \(H_0\) and find that \(\hat{\beta}\), the average causal effect of gender on salaries, is \(-1.069\). This means that on average, over the entire population, the odds of women having a salary \(>\$ 50K\) a year is \(\exp (-1.069) \approx 0.34\) times that of men, suggesting gender-based discrimination against women as measured by FACE. This finding is in agreement with the conclusions reported in \cite{nabi2018fair,li2017discrimination}. 
In the case of NYCSF data, we reject \(H_0\) and find that \(\hat{\beta}\) is \(0.273\) which means that on average, the odds of Black-Hispanics being arrested after a stop by the police, is \(\exp(0.273) \approx 1.31\) times that of Whites, suggesting possible racial bias against non-Whites.

\begin{table}[h]
	\caption{Estimates of FACE (\(\hat{\beta}\)) obtained on the synthetic, Adult, and NYCSF data sets.}
	\begin{tabular}{lccc}
		\toprule
		Data set   & \(\hat{\beta}\) & Standard Error & P-value \\
		\midrule
		Synthetic & \(-1.130 \times 10^{-16}\)   & \(6.991 \times 10^{-17}\) & \(\mathbf{0.106}\)   \\
		Adult     & \(\mathbf{-1.069}\)   & \(3.614 \times 10^{-1}\) & \(0.003\)  \\
		NYCSF & \(\mathbf{0.273}\)   & \(1.259 \times 10^{-1}\) & \(0.030\)  \\
		\bottomrule
	\end{tabular}
	\label{table:ace}
\end{table}

\subsection{FACT Check: Fairness Analysis Using FACT}
We report our analysis of fairness using FACT, for the synthetic, Adult, and NYCSF data sets.

\subsubsection*{{\bf Matching Quality Analyses}}
Because the quality of matched pairs used to estimate FACT impacts the conclusions that can be drawn using it, we compare the FACT estimates obtained using several widely-used matching methods described in Section \ref{matching_algo}. We present some analyses to verify that the generated matches are of sufficiently high quality for estimating FACT.

We observe that before matching, \(\overline{D}_{a, a^\prime}\) is 1.6400, 3.3508, and 1.1616, on the synthetic, Adult, and NYCSF data sets, respectively. The matching methods dramatically reduced \(\overline{D}_{a, a^\prime}\) on all of the data sets (see Table \ref{table:act}). Overall, NNM and FM achieved the lowest \(\overline{D}^{m}_{a, a^\prime}\) as compared to other matching methods on all data sets. The greater the number of pairs that are matched, the harder it is to achieve balance, and the trade-off between the two can be application dependent. We observed that considering the trade-off between the number of matches and \(\overline{D}^{m}_{a, a^\prime}\), FM yields higher quality matches on all data sets as compared to other methods. 

\begin{figure}[h]
	\hspace*{-0.5in}
	\centering
	\includegraphics[height = 5.6cm, width=\linewidth]{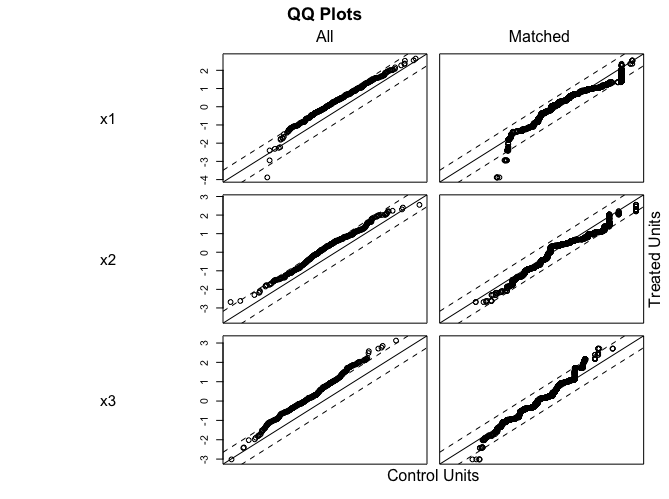}
	\caption{QQ plots of the first three features from the synthetic data set before (left) and after (right) FM.}
	\label{QQ}
\end{figure}

\begin{figure*}[t]
	\centering
	\begin{subfigure}{0.33\linewidth}        
		\centering
		\includegraphics[height=4.5cm,width=\linewidth]{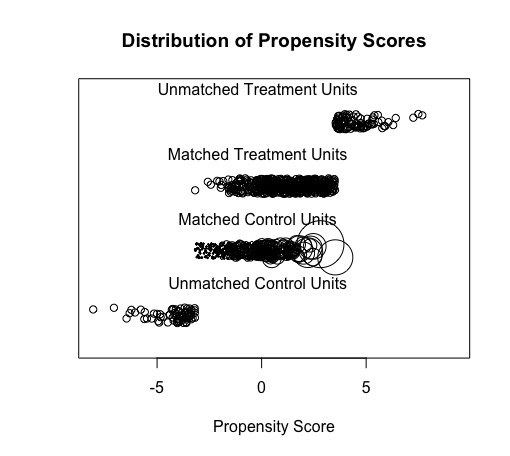}
		\caption{Synthetic data set.}
		\label{fig:A}
	\end{subfigure}
	\begin{subfigure}{0.33\linewidth}        
		\centering
		\includegraphics[height=4.5cm,width=\linewidth]{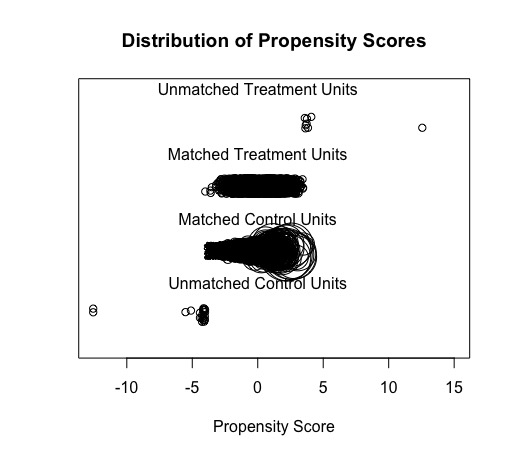}
		\caption{NYCSF data set.}
		\label{fig:A}
	\end{subfigure}
	\begin{subfigure}{0.33\linewidth}        
		\centering
		\includegraphics[height=4.5cm,width=\linewidth]{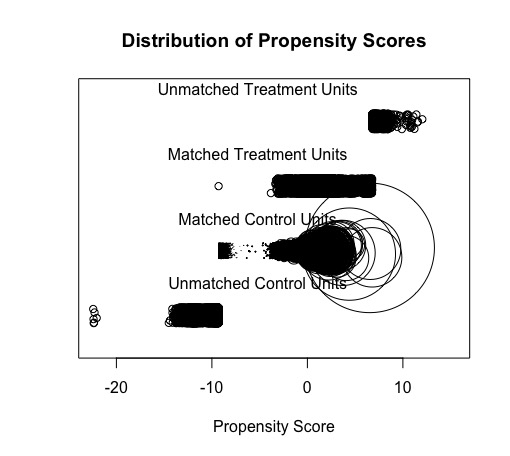}
		\caption{Adult data set.}
		\label{fig:B}
	\end{subfigure}
	\caption{Jitter plots of distribution of the propensity scores on the linear logit scale after FM on the synthetic (left), NYCSF (middle), and Adult (right) data sets. Each circle represents a data point. Area of the circle is proportional to the weight given to the data point. Female, Black-Hispanic = treated, and male, White = controlled.}
	\label{fig:jitter-all}
\end{figure*}

The QQ plots are generated for each feature in each data set. In Figure \ref{QQ} we show the QQ plots before and after FM for the first three features of the synthetic data set. The features lie far away from the 45 degree line before FM. After FM, the features are much better aligned to the diagonal line showing a more desirable feature balance. We also show the jitter plots of FM on all data sets in Figure 
\ref{fig:jitter-all}. It is clear that the distribution of propensity scores of the treated and controlled data points are very similar to each other after matching. Having verified that the results of matching are of adequate quality, we proceed to use them for estimating FACT.

\begin{table*}[t]
	\caption{Estimates of FACT (\(\hat{\gamma}\)) obtained via various matching methods on the synthetic, NYCSF, and Adult data sets.}
	\begin{tabular}{lcccccc}
		\toprule
		Synthetic data set &  &  &  &    &    &\\
		\midrule
		Matching Method &   \# of Treated Matches  &  \# of Control Matches   &    \(\overline{D}^{m}_{a, a^\prime}\)   &   \(\hat{\gamma}\)  &   Standard Error  &  P-value  \\
		\midrule
		NNM          & $ 454    $    & $ 155 $   & $ 0.0032 $    &  $-6.972 \times 10^{-17}$   &  $4.947 \times 10^{-17}$      & $ 0.159  $      \\
		NNMPC    & $ 454 $   & $ 168 $    & $ 0.0234 $ & $ -9.196 \times 10^{-17}$   & $ 3.020 \times 10^{-17}$ & $ 0.002  $   \\
		MMMPC & $ 454 $     & $ 174  $  & $ 0.0308 $ & $ 7.424 \times 10^{-18}$   & $ 3.192 \times 10^{-17}  $    & $ 0.816  $      \\
		FM              & $ 454 $    & $ 386 $    & $ 0.0031 $ & $ -8.263 \times 10^{-17} $   & $ 3.702 \times 10^{-17}  $   & $ 0.026 $      \\
		\toprule
		Adult data set &  &   &  &  &   &   \\
		\midrule
		Matching Method & \# of Treated Matches  & \# of Control Matches &  \(\overline{D}^{m}_{a, a^\prime}\)  &   \(\hat{\gamma}\)  &   Standard Error  &  P-value       \\
		\midrule
		NNM                 &$  13330   $   & $ 4922  $    & $ 0.0009  $   & $ -0.637 $      & $ 0.128 $   & $  6.560 \times 10^{-7} $ \\
		NNMPC    & $ 13301 $   & $ 5258 $    & $ 0.0714 $     & $ -0.650 $      &$  0.113  $ & $ 1.050 \times 10^{-8} $   \\
		MMMPC & $ 13301 $    & $ 5838 $    & $ 0.0584  $    & $ -0.586 $     & $ 0.131  $   & $ 7.650 \times 10^{-6} $  \\
		FM               & $ 13330  $     & $ 15320  $    & $ 0.0009  $   & $ -0.573  $     & $ 0.115 $    & $ 5.700 \times 10^{-7} $  \\
        \toprule
		NYCSF data set &  &   &  &  &   &  \\
		\midrule
		Matching Method &   \# of Treated Matches  &  \# of Control Matches   &    \(\overline{D}^{m}_{a, a^\prime}\)   &   \(\hat{\gamma}\)  &   Standard Error  &  P-value  \\
		\midrule
		NNM               & $ 2605  $     & $ 1305 $    & $ 0.0001  $   & $ 0.049  $     & $ 0.186 $    & $ 0.788 $ \\
		NNMPC               & $ 2605  $     & $1414 $    & $ 0.0266 $   & $ 0.246  $     & $ 0.160 $    & $ 0.124 $ \\
		MMMPC               & $ 2605  $     & $1264 $    & $ 0.0231 $   & $ 0.324 $     & $ 0.183 $    & $ 0.078 $ \\
		FM               & $ 2605  $     & $ 4958 $    & $ 0.0000  $   & $ 0.155  $     & $ 0.171 $    & $ 0.364 $  \\
		\bottomrule   
	\end{tabular}
	\label{table:act}
\end{table*}

\subsubsection*{{\bf FACT Estimates}}
The results of FACT analyses on the synthetic, Adult, and NYCSF data sets are summarized in Table \ref{table:act} (Note that EM did not yield any matches and hence is omitted from Table \ref{table:act}). In all cases, the null hypothesis is \(H_0:\gamma = 0\). In the case of synthetic data, FACT analyses show that for NNM and MMMPC, 
there is not enough evidence to reject \(H_0\). The p-values in the case of NNMPC and FM are \(<0.05\), but the magnitude of the estimated \(\hat{\gamma}\) is close to zero. We conclude that the synthetic data set is fair on average with respect to FACT. On the Adult data, we can reject \(H_0\), suggesting that salaries of women are significantly lower than those of men who match them on the non-protected attributes. For example, using FM, we find that \(\hat{\gamma} = -0.573\), thus the odds of women earning \(>\$50K\) a year, is \(\exp(-0.573) \approx 0.56\) times that of men. We conclude that in the Adult data, there is evidence of gender-based discrimination in salary, on average, against women. On the NYCSF data, interestingly, FACT analyses show that \(H_0\) cannot be rejected, suggesting a lack of evidence for racial bias, on average, in arrests after stops (when Black-Hispanics are compared with Whites who match them on non-protected attributes). 
This conclusion contradicts the finding of racial bias based on counterfactual fairness analysis (Supplementary Material S6 in \cite{kusner2017counterfactual}) which suggests discrimination against \emph{individuals}, as well as FACE analysis (see Section \ref{face}). We conjecture that the apparent discrepancy can be explained by noting that (i) fairness (or discrimination) on average does not necessarily imply individual-level fairness (or individual-level discrimination), and (ii) FACT compares the observed outcomes of members of a protected group with the hypothetical (counterfactual) outcomes they would have experienced had they not been members of the protected group (with all non-protected attributes remaining unchanged), whereas FACE compares such counterfactual outcomes on the entire population.

\subsubsection*{{\bf Impact of Unmeasured Confounders}}
We ran sensitivity analysis of our estimates of FACT for \(\Gamma = 1 ,\ldots, 10\) (where larger values of  \(\Gamma\) correspond to greater bias introduced by hidden confounders) on the Adult and NYCSF data sets. We find that all of our estimates obtained with various matching methods are quite robust to hidden confounder bias. Specifically, on the Adult data set, for all matching methods except FM, the estimates are robust to  such bias, and for FM, they are robust up to \(\Gamma = 4.5\), which corresponds to a fairly large amount of bias. On the NYCSF data set, estimates obtained via NNM and MMMPC are robust to hidden confounder bias, and NNMPC and FM are robust up to \(\Gamma\) equals 8.5, and 3, respectively. These results mean that our FACT estimates (and hence our findings of discrimination on the basis of protected attributes, or lack thereof) are fairly robust to hidden confounder bias.

\section{Summary and Discussion}
We have approached the problem of detecting whether a group of individuals that share a sensitive attribute, e.g., race, gender, have been subjected to discrimination in an algorithmic decision-making system, through the lens of causality. We have introduced two explicitly causal definitions of group fairness: {\em fair on average causal effect} (FACE), and {\em fair on average causal effect on the treated} (FACT). We have shown how to robustly estimate FACE and FACT, and use the resulting estimates to detect and quantify discrimination based on specific attributes (e.g., gender, race). The results of our experiments on synthetic data show that our proposed methods are effective at detecting and quantifying group fairness. Our analyses of the Adult data set for evidence of gender-based discrimination in salary, and of the NYCSF data set for evidence of racial bias in arrests after traffic stops, yield evidence of discrimination, or lack thereof, that is often in agreement with other studies.\footnote{The regression and matching-based methods we employed to estimate FACE and FACT adjust for covariates that might be potential confounders of the protected attribute, which although necessary in general, may be unnecessary in the case of gender and race, because they are unlikely to be caused by any other covariate. Consequently, the reported estimates of FACE and FACT are likely to represent {\em direct} causal effects as opposed to {\em total} causal effects.} We show on the real-world data that our estimates of FACE and FACT are robust to unmeasured confounding. Our results further show on the real-world data that FACE and FACT based findings do not always agree. Our FACT analyses also demonstrate that group-fairness (or discrimination) does not necessarily imply individual-level fairness (or individual-level discrimination).

Some directions for further research include: relaxing the assumption that the data are independent and identically distributed (i.i.d.) in settings where individuals are related to each other through family ties or other relationships; examining the relationships between different causal notions of fairness; and designing automated decision support systems that are demonstrably non-discriminatory with respect to given outcome(s) and protected attribute(s). 
\subsubsection*{{\bf Acknowledgements}} This work was funded in part by grants from the  NIH NCATS through the grant UL1 TR000127 and TR002014 and by the NSF through the grants 1518732, 1640834, and 1636795, the Edward Frymoyer Endowed Professorship in Information Sciences and Technology at Pennsylvania State University and the Sudha Murty Distinguished Visiting Chair in Neurocomputing and Data Science funded by the Pratiksha Trust at the Indian Institute of Science (both held by Vasant Honavar). The content is solely the responsibility of the authors and does not necessarily represent the official views of the sponsors.
\newpage

\bibliographystyle{ACM-Reference-Format}
\balance
\bibliography{bibliography}

\end{document}